\documentclass[letterpaper, 10 pt, conference]{ieeeconf}  

\IEEEoverridecommandlockouts                              

\usepackage{amsmath} 
\usepackage{amssymb}  
\usepackage{subcaption}
\usepackage{graphicx}
\usepackage{multirow}
\usepackage{tabularx}
\usepackage{caption}
\usepackage{subcaption}
\usepackage{adjustbox}
\usepackage{hyperref} 
\usepackage{color}
\usepackage{xcolor}

\usepackage[normalem]{ulem} 

\title{\LARGE \bf
A Variational Graph Autoencoder for \\  Manipulation Action Recognition and Prediction
}

\author{Gamze Akyol$^{1}$, Sanem Sariel$^{1}$, Eren Erdal Aksoy$^{2}$
\thanks{*This research is supported by a grant from the Scientific and Technological Research Council of Turkey (TUBITAK), Grant No. 119E-436}
\thanks{$^{1}$Artificial Intelligence and Robotics Laboratory, Faculty of Computer and Informatics Engineering, Istanbul Technical University, Maslak, Turkey
      {\tt\small akyolga@itu.edu.tr, sariel@itu.edu.tr}}
\thanks{$^{2}$School of Information Technology, Center for Applied Intel-
ligent Systems Research, Halmstad University, Halmstad, Sweden
        {\tt\small eren.aksoy@hh.se}}%
}
\begin{document}

\maketitle
\thispagestyle{empty}
\pagestyle{empty}

\begin{abstract}
Despite decades of research, understanding human manipulation activities is, and has always been, one of the most attractive and challenging research topics in computer vision and robotics. Recognition and prediction of observed human manipulation actions have their roots in the  applications related to, for instance, human-robot interaction and robot learning from demonstration. The current research trend heavily relies on advanced convolutional neural networks to process the structured Euclidean data, such as RGB camera images. These networks, however, come with immense computational complexity to be able to process high dimensional raw data.

Different from the related works, we here introduce a deep  graph autoencoder to jointly learn  recognition and prediction of manipulation  tasks from  symbolic scene graphs, instead of relying on the structured Euclidean data. Our network has a variational autoencoder structure with two branches: one for identifying the input graph type and one for predicting the future graphs. The input of the proposed network is a set of semantic  graphs which store the spatial relations between subjects and objects in the scene. The network output is a label set representing the detected and predicted class types. We benchmark our new model against different state-of-the-art methods on two different datasets, MANIAC and MSRC-9, and show that our proposed model can achieve better performance. {We also release our source code \href{https://github.com/gamzeakyol/GNet}{https://github.com/gamzeakyol/GNet}.}
\end{abstract}

\section{Introduction}

Manipulation action recognition is an important cognitive faculty and has numerous  potential applications in   computer vision and robotics. Video surveillance, human-robot interaction, and robot learning from demonstration are to name a few. There exist various advanced  network architectures \cite{donahue2015long, feichtenhofer2016convolutional, ji20123d, zhou2018temporal} exhibiting promising performances in the task of action identification. Such networks, however,   work only with the structured Euclidean data, and   come with immense computational complexity to be able to process high dimensional raw data.
  
There is already a large corpus of work \cite{aksoyIJCV, chen2019graph, dreher2019learning, herzig2019spatio, huang2020spatio, ji2020action, materzynska2020something, wang2018videos} showing that employing the symbolic graph representation of the   scene as an input can lead to a better action recognition performance than networks of using the structured Euclidean data streams, e.g. RGB images. This is mainly because symbolic graphs provide compact and descriptive expression  of the scene. In this regard, Graph Neural Networks \cite{scarselli2008graph} have also become \textit{de facto} standards due to their flexibilities in processing non-Euclidean data streamed in different sizes.  

With this motivation, we here introduce a variational deep graph autoencoder receiving the scene graph  as  input not only to recognize  the input graph type but also to predict the types of future graphs of the given manipulation. 
Our proposed graph network has an encoder-decoder structure, where encoder involves  multiple graph convolution~\cite{morris2019weisfeiler} layers and decoder has two branches: one is for the action recognition task and the other is for prediction. 
In the recognition branch, \textit{linear} and \textit{softmax} layers are introduced, whereas the action prediction branch involves graph recurrent layers (Set2Set structure~\cite{vinyals2015order}) to  learn temporal features between input graphs.
Our input scene graphs capture semantic spatial relations (such as touching) between subjects and objects in the scene to form the descriptive symbolic scene representation. For instance, image segments form graph nodes while edges define spatial relations as in \cite{aksoyIJCV}. The network output is a set of labels indicating the detected and   predicted  manipulation types. In contrast to previous related studies \cite{dreher2019learning, kipf2016variational}, our graph network differs in that it performs a multitask learning (i.e., graph recognition and prediction) using a single and compact network representation.

We apply our model to the manipulation action recognition and prediction tasks using a publicly available MANIAC dataset~\cite{aksoy2015model}. We additionally compare the performance of our graph network with the other strong baselines \cite{neumann2016propagation, neumann2012efficient} using the MSRC-9~\cite{winn2005object} dataset.  Experimental findings reported here show that  our model outperforms the others in the task of manipulation action recognition. 

Organization of the paper structure is as follows: related works in manipulation action recognition and prediction literature are presented in section 2 together with the graph neural networks literature. In section 3, our network architecture is presented in detail with the input graph representation. In the next section, experimental results are given on datasets in addition to implementation details. Section 5 presents conclusion and future directions.

\section{Related Work}

We first review the literature on action recognition and prediction. 
We further examine the multi-task learning with graph neural networks as we aim to execute both recognition and prediction tasks using a single graph neural network model. 

Note that \textit{action classification} and \textit{action recognition} terms are used interchangeably in the literature. In our study, we  follow the   taxonomy in \cite{hutchinson2020video} and use  the term \textit{action classification} as a general term for action recognition. 

\textbf{Manipulation Action Recognition and Prediction:}
Action recognition is a classification task of a given complete observation sequence \cite{cheng2015advances, herath2017going, hutchinson2020video, kong2018human}. This problem is a sequential data classification problem which includes also spatial processing.  Thus, it becomes a challenging spatial-temporal problem.

There are deep learning based space-time networks \cite{ji20123d}, multi-stream networks \cite{feichtenhofer2016convolutional}, and hybrid networks \cite{donahue2015long}.
The space-time network in    \cite{ji20123d}   performs 3D convolutions over adjacent frames to extract features from both spatial and temporal dimensions. 
Multi-stream networks  solve the aggregation of temporal information by introducing a temporal stream. For instance, the work in \cite{feichtenhofer2016convolutional} uses two stream convolutional networks: spatial stream holds static frames and temporal stream holds the changing frames to capture motion cues.
Hybrid networks use recurrent layers on convolutional networks to solve the temporal data aggregation. The hybrid network in \cite{donahue2015long} utilizes long-short term memory (LSTM) layers on  convolutional networks to obtain temporal information while the convolutional part captures the spatial cues. 

Yang et. al. \cite{yang2019learning} present  a manipulation  recognition model using two consecutive neural networks to estimate the action labels from human demonstration videos. The first network is an object detection model using raw RGB images from the video, whereas the subsequent network  estimates the action types relying on these detected objects  and raw images. Authors use the first and last frames of the video as the input  to simplify the problem by eliminating irrelevant data. However, such a frame dropping process without considering the relevance can lead to certain information loss.

Dreher et. al. \cite{dreher2019learning} propose a graph encoder-decoder network to recognize human demonstrated bimanual manipulation actions. This network uses objects and their relations encoded as scene graphs similar to our work. Our network differs in that we solve the action recognition and prediction tasks jointly using a single network which is not the case in  \cite{dreher2019learning}.

In the case of action prediction, the main goal is rather to classify actions from incomplete input sequence. According to taxonomy introduced in \cite{hutchinson2020video}, there exist two kinds of action prediction methods: action anticipation and early action prediction.  Action anticipation is a prediction problem from an unobserved sequence. Early action prediction is, on the other hand, doing prediction from  early frames coming from an image sequence. In this manner, our work falls into the  early action prediction category.

In \cite{yu2020novel}, a hybrid deep neural network is used for the action prediction. They use convolution and LSTM layers in order to solve the action prediction problem only. 

\textbf{Graph Neural Networks:}
Geometric deep learning and graph neural networks (GNNs) are being to be popular in various domains due to their flexibility in working with non-Euclidean data and different sized inputs.  
Standard deep learning methods can, however, work only with Euclidean structured data such as images and sequential data. 

There are four main subcategories for GNNs in the literature \cite{wu2020comprehensive}, which are: Graph Convolutional Neural Networks \cite{kipf2016semi}, Recurrent Graph Neural Networks \cite{vinyals2015order}, Spatial-temporal Graph Neural Networks \cite{huang2020spatio}, and Graph Autoencoders \cite{kipf2016variational}. By employing these networks, graphs can be classified into node-level, edge-level or graph-levels.
For instance, node-level classification is predicting labels, i.e. attributes, of the nodes in a graph \cite{dabhi2020nodenet, feng2020graph, velivckovic2017graph}. 
Edge-level classification is rather predicting labels of edges in attributed graphs \cite{gu2019link, kipf2016variational, pan2018adversarially}. 
Graph-level classification is trying to predict labels of the input  graphs \cite{chauhan2020few, chen2020convolutional, errica2019fair, sun2019infograph}. 
In our proposed work, we use graph autoencoders to  handle action recognition and prediction tasks as a graph-level classification problem.

Graph autoencoders \cite{kipf2016variational, simonovsky2018graphvae, tran2018multi} are unsupervised learning methods which encode the graph with an encoder and perform inferences from a latent space using a decoder. This method can be employed in a graph reconstruction task \cite{tran2018multi}. In \cite{kipf2016variational}, a variational graph autoencoder model  is proposed to solve the edge prediction problem. This model can learn latent representations for undirected graphs in an unsupervised manner.
Multi-task graph autoencoder in \cite{tran2018multi} is proposed  for addressing the edge prediction and node classification problems.  This model shares parameters between the encoder and decoder parts of the autoencoder, which boosts the model performance.
In \cite{simonovsky2018graphvae}, the authors propose a variational autoencoder to learn from continuous embedding for the graphs.
Although this model can learn small graph embedding, it has difficulty with large graphs.

In this work, we use variational graph autoencoder since the latent space representation can help to solve multiple learning tasks simultaneously. 
More specifically, we propose a compact graph neural network model capable of performing both action recognition and prediction which is the   core difference from previous works in the literature.

\section{Method}
In this section, we introduce our network model to solve the action recognition and prediction problems.
We leverage graph neural network layers for developing our network architecture. More specifically, we use a graph autoencoder architecture based on the variational graph autoencoder (VGAE) model introduced in~\cite{kipf2016variational}. The base VGAE architecture is particularly designed to deal with the graph edge prediction problem only. In our case, we extend VGAE to jointly perform both action recognition and prediction, each of which was treated as a graph classification problem. The input is a graph representing, for instance, the observed scene. The output is a set of labels representing the detected and predicted actions. In the following subsections, we provide detailed descriptions of the input graph  structure  and the network architecture. 

\subsection{Input Graph Representation}
Our network uses scene graphs as the input. A scene graph $\mathcal{G}$ is formulated as $\mathcal{G}=(\mathcal{V}, \mathcal{E})$ where $\mathcal{|V|} = N$, $v_i \in \mathcal{V}$, and $\mathcal{|E|} = M$ with $e_{i,j} \in \mathcal{E}$, $ v_i \in \mathcal{R}^{d_v}$, and $ e_{i,j} \in \mathcal{R}^{d_e}$.
Here, $\mathcal{V}$ and $\mathcal{E}$  represent nodes and edges of the graph. $N$ is the number of the nodes and $M$ is the total edge number.

\begin{figure}[!t]
\centering
\begin{subfigure}{.46\columnwidth}
  \centering
  \includegraphics[width=\columnwidth, keepaspectratio]{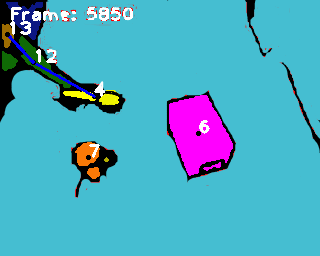}
  \label{fig:key1a}
\end{subfigure}
\begin{subfigure}{.46\columnwidth}
  \centering
  \includegraphics[width=\columnwidth, keepaspectratio]{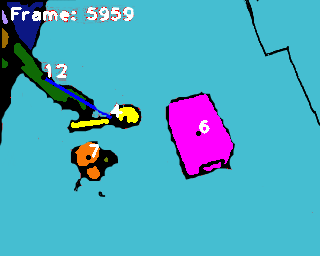}
  \label{fig:key1b}
\end{subfigure}

\begin{subfigure}{.46\columnwidth}
  \includegraphics[width=\columnwidth, keepaspectratio]{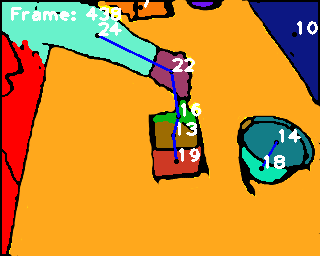}
  \label{fig:key2a}
\end{subfigure}
\begin{subfigure}{.46\columnwidth}
  \includegraphics[width=\columnwidth, keepaspectratio]{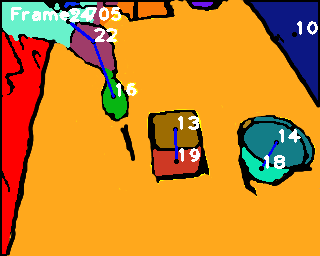}
  \label{fig:key2b}
\end{subfigure}
\caption{Sample frames from different action classes in the MANIAC dataset \cite{aksoy2015model}. The top and bottom rows show the pushing and stirring actions, respectively. Each color here represents a unique image segment (i.e., an object), from which graphs are derived.  Relations between ``touching" segments are shown with the blue edges.}
\label{fig:keyframes}
\end{figure}

Following the work in \cite{aksoyIJCV}, the input camera image is semantically segmented to be represented by a scene graph in which nodes indicate the semantic segments (e.g., a unique color for each object) and edges represent the spatial contact relations (such as touching) between segments.
This yields a scene graph with an arbitrary number of nodes and edges to be fed to the network.
Feature representations of nodes are made up of object classes. The graph is an unweighted graph, so edge values are either 0 or 1.
Fig.~\ref{fig:keyframes} shows  sample scene graphs together with the corresponding semantic segments from the MANIAC dataset. 


\begin{figure*}[ht!]
\centering
\includegraphics[width=0.8\textwidth]{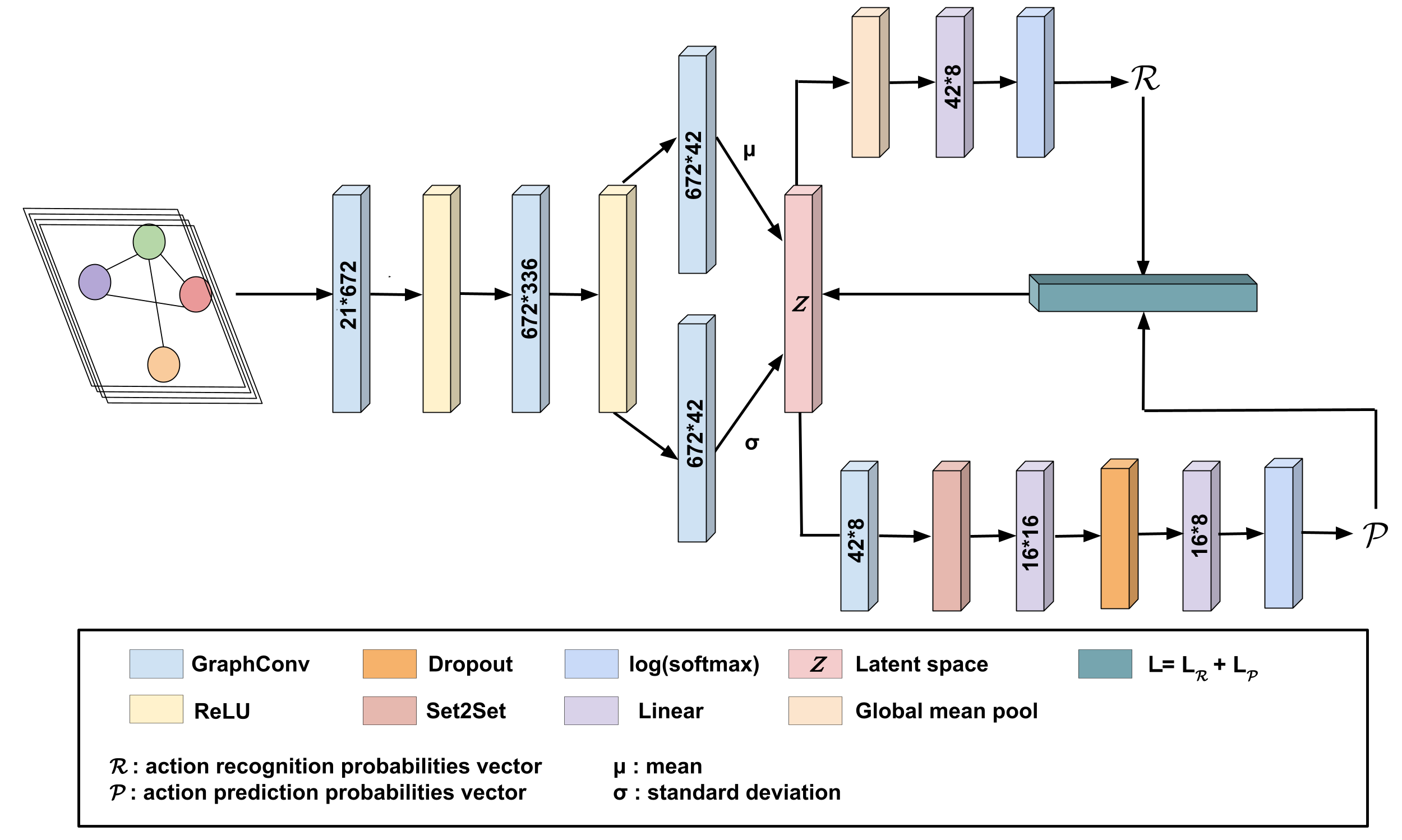}
\caption{Network architecture: Input is designed to support multiple graphs. Output involves the terms $\mathcal{R}$ and $\mathcal{P}$ representing action recognition and action prediction labels, respectively. Numbers indicate input and output sizes of layers, respectively. The final loss is computed using  both outputs $\mathcal{R}$ and $\mathcal{P}$.}
\label{network}
\end{figure*}
\subsection{Network Architecture}
The  architecture of our proposed   graph   network  is illustrated in Fig.~\ref{network}. The model has an encoder-decoder structure.   Encoder involves  $3$ graph convolution (GraphConv)~\cite{morris2019weisfeiler} layers and $2$ ReLUs.  
As the output of the third GraphConv layer, mean ($\mu$) and variance ($\sigma$) values are computed to return the latent vector $\mathbf{Z}$. 

GraphConv layers are used for learning the weights of the nodes and the relations between each node. Hidden layer representation of graph convolution layer is implemented according to the following equation in  \cite{morris2019weisfeiler}: 
\begin{equation}{
v_i^{'} = \theta_1v_i + \theta_2 \sum_{j\in N(i)}e_{j,i}\cdot v_j  \label{cnnformula}
}\end{equation}
In this formula, $v_i^{'}$  represents weight of the node $i$ in the next iteration. $v_i$ is current weight of the node  $i$ and $v_j$ defines current weight of the node $j$. $e_{j,i}$ represents the edge weight between the source node $j$ and the target node $i$. All edge weights are summed between the node $i$ and its neighbor nodes by the rate of node weights of neighbors. Here, $N(i)$ represents neighbours of the node $i$. $\theta_1$ and $\theta_2$ are the coefficients to be learned during training.

On the decoder side, we introduce two branches: one is for the action recognition task and the other is for prediction. 
For the recognition branch, \textit{linear} and \textit{softmax} layers are introduced after having a global mean pooling layer.
In the action prediction branch, graph recurrent layers (Set2Set structure~\cite{vinyals2015order}) are used for learning temporal features of graphs between different time steps.
LSTM \cite{hochreiter1997long} layers are further employed in the Set2Set module to increase the role of temporal dependencies between consecutive graphs. Finally, \textit{linear} and \textit{softmax}  layers are injected as in the case of the recognition branch. 
Hidden layer representation of the graph recurrent layer is implemented using the following  set of equations: 
\begin{equation*}{
        \mathbf{q}_t = \mathrm{LSTM}(\mathbf{q}^{*}_{t-1})
}\end{equation*}
\begin{equation*}{
        \alpha_{i,t} = \mathrm{softmax}(\mathbf{v}_i \cdot \mathbf{q}_t)
}\end{equation*}
\begin{equation*}{
        \mathbf{r}_t = \sum_{i=1}^N \alpha_{i,t} \mathbf{v}_i
}\end{equation*}
\begin{equation}{
        \mathbf{q}^{*}_t = \mathbf{q}_t \, \Vert \, \mathbf{r}_t \label{rnnformula}
}\end{equation}

where $\mathbf{q}_t$ represents the current weight computed from the weight coming from the previous time step, $\mathbf{q}^{*}_{t-1}$. $\mathbf{v}_i$ represents the node features of node $i$. $ \Vert \ $ is the concatenation operator which causes the output to double in size.

The total loss function ($\mathcal{L}$) is formulated as a linear combination of both action recognition ($\mathcal{L_{R}}$) and prediction ($\mathcal{L_{P}}$)  losses, each of which has a negative log likelihood:
\begin{equation}
    \mathcal{L} = \mathcal{L_{R}} + \mathcal{L_{P}}  \label{loss}
\end{equation}

Adam optimizer \cite{kingma2014adam} is used for the optimization with the learning rate of $1e-6$. The batch size and dropout probability values are set to $1$ and $0.5$, respectively.  


\section{Experimental Results}

\subsection{Datasets}
MANIAC dataset \cite{aksoy2015model} is used for action recognition and   prediction tasks. Microsoft Research Challenge (MSRC-9) graph dataset \cite{winn2005object} is   used for comparing our method with other existing methods in the task of action recognition only.

\textbf{MANIAC Dataset:}
This dataset \cite{aksoy2015model} already comes with the scene graph sequences of videos where subjects demonstrate $8$ different manipulation actions which are: \textit{ chopping, cutting, hiding, pushing, putting on top, stirring, taking down,} and \textit{uncovering}. These action names are employed as graph labels for the action recognition and prediction tasks. Each action contains $15$ different demonstrations, i.e. there exist overall 120 action samples. 
In all these action videos, in total, $30$ different objects are manipulated from $14$ unique object classes. 
In the MANIAC videos, each frame is represented by a graph where nodes hold object type information. Edges are established between nodes to define their pairwise spatial object relations, such  as ``touching" or ``non-touching". 
Sample graphs from the MANIAC dataset can be seen in Fig.~\ref{fig:keyframes}. For instance, in the top row in Fig.~\ref{fig:keyframes}, the graph nodes $4$, $6$, $7$ and $12$ are the object labels. Note that there emerges no edge between nodes $6$ and $7$, since objects are well separated from each other.  In contrast, various edges are constructed between graph nodes in the second row in Fig.~\ref{fig:keyframes} as most objects are touching each other.

\begin{figure}[!b]
\centering
\begin{subfigure}{.46\columnwidth}
  \centering
  \includegraphics[width=\columnwidth]{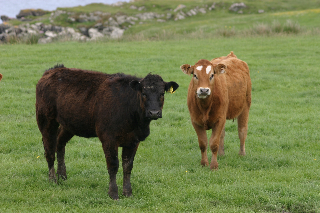}
  \label{fig:key1l}
\end{subfigure}
\begin{subfigure}{.46\columnwidth}
  \centering
  \includegraphics[width=\columnwidth]{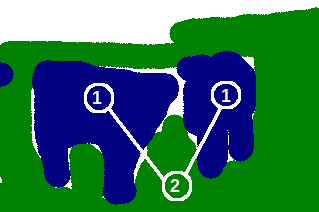}
  \label{fig:key1r}
\end{subfigure}

\begin{subfigure}{.46\columnwidth}
  \includegraphics[width=\columnwidth]{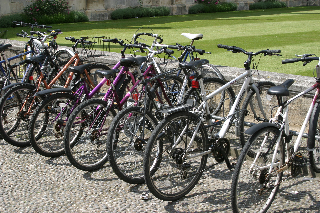}
  \label{fig:key2l}
\end{subfigure}
\begin{subfigure}{.46\columnwidth}
  \includegraphics[width=\columnwidth]{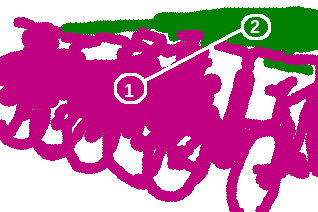}
  \label{fig:key2r}
\end{subfigure}
\caption{Sample images from different classes in the MSRC-9 dataset \cite{winn2005object}. Left column shows sample raw images from \textit{cow} and \textit{bike} classes, respectively. Right column shows their segmented versions together with the constructed scene graphs. }
\label{fig:msrc9samples}
\end{figure}

\textbf{MSRC-9 Dataset:}
This dataset \cite{winn2005object} contains $240$ real-world images hand labeled with $9$ classes: \textit{ building, grass, tree, cow, sky, plane, face, car,} and \textit{bike}. When converting images in this dataset into  graphs, some images are discarded by \cite{neumann2016propagation}. Consequently, there are in total $221$ graphs from $8$ unique classes. Graph nodes are image superpixels and edges are constructed between neighbour superpixels.
Fig. \ref{fig:msrc9samples} shows sample images and generated graphs from the MSRC-9 dataset.

\subsection{Implementation Details}

Train, validation, and test set split ratios are $10$,  $3$, and $2$ for the MANIAC dataset and  $8$, $1$, and $1$ for MSRC-9.
In the MANIAC dataset, $4$ consecutive scene graphs are merged into one to be fed to the network as an input, whereas graphs are treated individually in MSRC-9 to have a fair comparison with other state-of-the-art models \cite{neumann2016propagation, neumann2012efficient}. 
In the MANIAC dataset, input size for the first layer of the encoder is set to $21$ representing the number of node features. Output channel size for the first layer is set to $672$. 
For MSRC-9 dataset, input size for the first layer of the encoder is a number of node features again, which is 10. Channel size for the output of the first layer is 1280.
Note that different channel sizes have been chosen to enhance learning, since the number of data and node features vastly vary between datasets.
Our proposed network is trained for $200$ and $500$ epochs for the MANIAC and MSRC-9 datasets, respectively.
We  release the code for public use (\href{https://github.com/gamzeakyol/GNet}{https://github.com/gamzeakyol/GNet}.)

\subsection{Quantitative Results}

We first evaluate the performance of our network on the MANIAC dataset. Table \ref{tab:results1} reports the obtained quantitative results for the validation and test splits when action recognition and prediction branches are trained  individually and also together in a joint mode. The last two rows in this table clearly show that training both branches together leads to better accuracy scores in the test splits. This is because forcing the network to jointly learn both recognition and prediction tasks helps prevent the overfitting problem. Note that unlike the second and third rows showing the individual performances of  branches, the difference between the validation and test scores are getting reduced in the last two rows. 

In Table \ref{tab:results1}, we also compare our results with the Semantic Event Chain (SEC) based action recognition method introduced in \cite{aksoy2015model}. We here note that to solve the same recognition task, the SEC method \cite{aksoy2015model} utilizes the entire graph sequence in one manipulation demonstration and also treats the \textit{cutting} and \textit{chopping} actions as the same. Therefore, it may not be fair to compare with our method, although our proposed model exhibits better recognition performance. 

\begin{table}[t!]
\centering
\caption{Quantitative results on the MANIAC dataset. Validation and test accuracy scores are given in percentage (\%).  \label{tab:results1}}
\scalebox{1.25}{
\begin{tabular}{|l|l|l|l|}
\hline
\multicolumn{1}{|l}{} & \multicolumn{1}{l|}{} & Val & Test \\ \hline
\multicolumn{2}{|l}{SEC-based Recognition  \cite{aksoy2015model} } & \multicolumn{1}{|l|}{-} & 75.87 \\ \hline
\multicolumn{2}{|l}{GNet-Recognition Only} & \multicolumn{1}{|l|}{80.28} & 72.44 \\ \hline
\multicolumn{2}{|l}{ GNet-Prediction Only} & \multicolumn{1}{|l|}{64.08} & 59.38 \\ \hline
\multirow{2}{*}{\textbf{\rotatebox[origin=c]{0}{GNet}}} &
Recognition  & 77.77 & \textbf{77.56} \\ \cline{2-4}
& Prediction & 66.20 & 65.34 \\ \hline
\end{tabular}}
\end{table}
 
Next, we benchmark our proposed network on the MSRC-9 dataset. To have a fair comparison with the other related works \cite{neumann2016propagation, neumann2012efficient}, we restrict  this benchmarking to action recognition task only. Obtained results are reported in Table \ref{tab:baselines}. This table clearly shows  that our network considerably outperforms the others by leading to the highest recognition score ($95.4\%$) which is $3.3\%$ over the previous state-of-the-art method~\cite{neumann2012efficient}.

 \begin{table}[b!]
\centering
\caption{Action recognition results on the MSRC-9 dataset. Test accuracy scores are given in percentage (\%).   \label{tab:baselines}}
\scalebox{1.25}{
\begin{tabular}{|l|l|c|}
\hline
\multicolumn{1}{|l}{} & \multicolumn{1}{l|}{} & Test \\ \hline
\multirow{4}{*}{\textbf{\rotatebox[origin=c]{90}{Baselines}}}                       & PK \cite{neumann2016propagation} & 90.0 \\ 
 & $K_{DIFF+H}$ \cite{neumann2012efficient} & 91.6 \\ 
 & $K_{DIFF+TV}$ \cite{neumann2012efficient} & 91.6 \\
 & $K_{WL}$  \cite{neumann2012efficient} & 92.1 \\ \hline
\multicolumn{1}{|l}{} & \multicolumn{1}{l|}{GNet-Recognition Only} &  {\bf 95.4}  \\ \hline
\end{tabular}}
\end{table}

As baseline models, we use the state-of-the-art works in  \cite{neumann2016propagation} and \cite{neumann2012efficient}. 
Note that in Table \ref{tab:baselines}, PK stands for \textit{``Propagation Kernel"}~\cite{neumann2016propagation} where $20\%$ of labels are omitted. K stands for \textit{``Kernel"}\cite{neumann2012efficient} and uses the entire dataset. $K_{DIFF+H}$ represents diffusion graph kernel using Hellinger distance, and \textit{(+TV)} stands for total variation distance. $K_{WL}$ represents \textit{``WL-subtree kernel"}.
In \cite{neumann2012efficient}, the authors try different kernels using continuous vector valued node labels.
The work in \cite{neumann2016propagation} is a graph-kernel framework for calculating similarities of graphs by propagating continuous node features.
Also in the previous study \cite{neumann2012efficient}, Neumann et. al. introduce graph kernels for continuous node features instead of discrete node features, where different parametrized kernels are used for different experiments.
Thanks to our encoder-based model, the graph embeddings in the latent space leads to higher recognition performance in contrast to these baseline models.


\section{Conclusion}
In this work, we introduce a variational deep graph autoencoder model to jointly learn manipulation action recognition and prediction. We benchmark our new model against different state-of-the-art methods on two different datasets and show that our proposed model can achieve better performance. 

We plan to extend our model with additional node and edge prediction branches in order to generate the predicted action graphs.
We further plan to consider the temporal information in our network structure to boost the overall network performance.





{
\bibliographystyle{IEEEtran}
\bibliography{ref}
}

\end{document}